\documentclass[a4paper,conference]{IEEEtran}
\usepackage{cite}
\ifCLASSINFOpdf
  \usepackage[pdftex]{graphicx}
\else
\fi
\usepackage{amssymb}
\usepackage{amsmath}
\ifCLASSOPTIONcompsoc
\usepackage[caption=false,font=normalsize,labelfont=sf,textfont=sf]{subfig}
\else
  \usepackage[caption=false,font=footnotesize]{subfig}
\fi
\usepackage{url}
\usepackage{makecell}
\usepackage{flushend}
\begin{document}
%
% paper title
% Titles are generally capitalized except for words such as a, an, and, as,
% at, but, by, for, in, nor, of, on, or, the, to and up, which are usually
% not capitalized unless they are the first or last word of the title.
% Linebreaks \\ can be used within to get better formatting as desired.
% Do not put math or special symbols in the title.
\title{ASL Recognition with Metric-Learning based Lightweight Network}

% author names and affiliations
% use a multiple column layout for up to three different
% affiliations
\author{
\IEEEauthorblockN{Evgeny Izutov}
\IEEEauthorblockA{
IOTG Computer Vision (ICV), Intel\\
Email: evgeny.izutov@intel.com}
}

% conference papers do not typically use \thanks and this command
% is locked out in conference mode. If really needed, such as for
% the acknowledgment of grants, issue a \IEEEoverridecommandlockouts
% after \documentclass

% for over three affiliations, or if they all won't fit within the width
% of the page, use this alternative format:
%
%\author{\IEEEauthorblockN{Michael Shell\IEEEauthorrefmark{1},
%Homer Simpson\IEEEauthorrefmark{2},
%James Kirk\IEEEauthorrefmark{3},
%Montgomery Scott\IEEEauthorrefmark{3} and
%Eldon Tyrell\IEEEauthorrefmark{4}}
%\IEEEauthorblockA{\IEEEauthorrefmark{1}School of Electrical and Computer Engineering\\
%Georgia Institute of Technology,
%Atlanta, Georgia 30332--0250\\ Email: see http://www.michaelshell.org/contact.html}
%\IEEEauthorblockA{\IEEEauthorrefmark{2}Twentieth Century Fox, Springfield, USA\\
%Email: homer@thesimpsons.com}
%\IEEEauthorblockA{\IEEEauthorrefmark{3}Starfleet Academy, San Francisco, California 96678-2391\\
%Telephone: (800) 555--1212, Fax: (888) 555--1212}
%\IEEEauthorblockA{\IEEEauthorrefmark{4}Tyrell Inc., 123 Replicant Street, Los Angeles, California 90210--4321}}

% use for special paper notices
%\IEEEspecialpapernotice{(Invited Paper)}

% make the title area
\maketitle

% As a general rule, do not put math, special symbols or citations
% in the abstract
\begin{abstract}
In the past decades the set of human tasks that are solved by machines
was extended dramatically. From simple image classification problems researchers
now move towards solving more sophisticated and vital problems, like, autonomous
driving and language translation. The case of language translation includes a
challenging area of sign language translation that incorporates both image and
language processing. We make a step in that direction by proposing a lightweight
network for ASL gesture recognition with a performance sufficient for practical
applications. The proposed solution demonstrates impressive robustness on MS-ASL
dataset and in live mode for continuous sign gesture recognition
scenario. Additionally, we describe how to combine action recognition model
training with metric-learning to train the network on the database of limited
size. The training code is available as part of Intel$^{\textregistered}$ OpenVINO\texttrademark Training Extensions.

\end{abstract}

% no keywords

% For peer review papers, you can put extra information on the cover
% page as needed:
% \ifCLASSOPTIONpeerreview
% \begin{center} \bfseries EDICS Category: 3-BBND \end{center}
% \fi
%
% For peerreview papers, this IEEEtran command inserts a page break and
% creates the second title. It will be ignored for other modes.
\IEEEpeerreviewmaketitle

\section{Introduction}
Humanity put artificial intelligence into
service in a wide range of applied tasks. Nonetheless, for a number of problems
we are still trying to get closer to the human-level performance.  One of such
challenges is a sign language translation that can help to overcome the
communication barrier between larger number of groups of people.

There are millions of people around the world, who use one from over
several dozens of sign languages (e.g. ASL in United States and most of
Anglophone Canada, RSL in Russia and neighboring countries, CSL in China,
etc.). A sign language itself is a natural language that uses the visual-manual
modality to represent meaning through manual articulations. It goes without saying
that sign language is different from the common language in the same country by
its grammar and lexicon - it's not just a literal translation of single words in
a sentence. In addition, sign language from a certain country can have different
dialects in various locations. The latter aspect significantly complicates
solving the sign language recognition problem due to the need of a large and
diverse database.

To tackle this challenge, researchers have tried to use methods from the
adjacent action recognition area like 3D convolution networks
\cite{DBLP:journals/corr/TranBFTP14}, two-stream networks with additional depth
or flow stream \cite{DBLP:journals/corr/SimonyanZ14}, skeleton-based action
recognition \cite{hosain2019sign}. Unfortunately, the aforementioned approaches
don't work very well in case of petty size datasets that we are dealing with in
the sign language recognition space.  Another issue is related to the inference
speed - the network needs to run in real-time to be useful in live usage
scenarios.

To solve the listed problems we propose several architectural choices
(namely, applying the 2D depth-wise framework to 3D case) and a training
procedure that aims to combine a metric-learning paradigm with continuous-stream
action recognition. Summarizing all of the above, our contributions are as
follows:
\begin{itemize}
\item Extending the family of efficient 3D networks
for processing continuous video stream by merging S3D framework
\cite{DBLP:journals/corr/abs-1712-04851} with lightweight edge-oriented
MobileNet-V3 \cite{DBLP:journals/corr/abs-1905-02244} backbone architecture.
\item Introducing residual spatio-temporal attention module with auxiliary loss
to control the sharpness of the mask by using Gumbel sigmoid
\cite{jang2016categorical}.
\item Using metric-learning techniques to deal
with limited size of ASL datasets to reach robustness.
\end{itemize}

Finally, the model trained on the MS-ASL dataset
\cite{DBLP:journals/corr/abs-1812-01053} is prepared for inference by
Intel$^{\textregistered}$ OpenVINO\texttrademark
toolkit\footnote{\url{https://software.intel.com/en-us/openvino-toolkit}} and
is available as a part of the Intel$^{\textregistered}$ OpenVINO\texttrademark
OMZ\footnote{\url{https://github.com/opencv/open_model_zoo}}. There you can
find sample code on how to run the model in demo mode. In addition, we
release the training
framework\footnote{\url{https://github.com/opencv/openvino_training_extensions}}
that can be used in order to re-train or fine-tune our model with a custom database.

\section{Related Work}

\paragraph{Action Recognition}
Recent developments in deep learning helped to
make a step from well-studied image-level problems (e.g. classification,
detection, segmentation) to video-level problems (forecasting, action
recognition, temporal segmentation). Such domain difference appears by
introducing an extra temporal dimension. First solutions used direct
incorporation of motion information by processing motion fields in two-stream
network \cite{DBLP:journals/corr/SimonyanZ14}. Another approach was based on a
simple idea of extending of common 2D convolutions to 3D case: C3D
\cite{DBLP:journals/corr/TranBFTP14}, I3D \cite{DBLP:journals/corr/CarreiraZ17}.
The main disadvantage of aforementioned methods was the inability to train deep
3D networks from scratch because of over-fitting on target datasets (note that
collecting a dataset close to ImageNet by size and impact \cite{ILSVRC15} for
video-oriented problems is still challenging). Next steps were focused on
reducing the number of parameters and thereby decreasing over-fitting by using
separable 3D convolutions (P3D \cite{DBLP:journals/corr/abs-1711-10305} and
R(2+1)D \cite{DBLP:journals/corr/abs-1711-11248} networks) and investigating the
ability to mix 2D and 3D building blocks inside a backbone in an optimal way
\cite{DBLP:journals/corr/abs-1712-04851}. We follow the same concept as in S3D
\cite{DBLP:journals/corr/abs-1712-04851} but use depth-wise convolutions as in
the original 2D MobileNetV3 architecture
\cite{DBLP:journals/corr/abs-1905-02244}.

Other research directions are based on the ideas of using appearance from
appropriate (key) frames rather than any kind of motion information
\cite{DBLP:journals/corr/WangXWQLTG17}, to mix motion information on feature
level by shifting channels \cite{DBLP:journals/corr/abs-1811-08383}, to
incorporate relational reasoning over frames in videos
\cite{DBLP:journals/corr/abs-1711-08496}. Unfortunately, as it was shown in
\cite{DBLP:journals/corr/abs-1812-01053} the appearance- and late-fusion-
\cite{DBLP:journals/corr/abs-1804-09066} based methods are not able to recognize
quick gestures like sign language due to insufficient information at the
single-frame level. Most hand gestures are, essentially, a quick movement of
fingers and it's impossible to recognize it by inspecting any single image
from the video sequence -- it should be considered in full.

\paragraph{ASL Recognition}
Following the success of CNNs for action
recognition, the first sign language recognition approaches tried to reuse 3D
convolutions \cite{Pigou_2017_ICCV} to use frame-level \cite{Shi_2019_ICCV} or
skeleton \cite{DBLP:journals/corr/abs-1802-07584},
\cite{DBLP:journals/corr/abs-1901-11164}, \cite{hosain2019sign} feature fusion
by recurrent networks \cite{DBLP:journals/corr/abs-1808-03314} or graph
convolution networks \cite{DBLP:journals/corr/abs-1901-00596}. Other approaches
use multi-stream and multi-modal architectures to capture motion of each hand
and head independently \cite{yang2019sfnet}, mix depth and flow streams
\cite{DBLP:journals/corr/abs-1906-02851}.

Aforecited methods talk about sign level recognition problem rather than
sentence translation. To solve the translation problem, another kind of language
model is trained: \cite{Pu_2019_CVPR}, \cite{DBLP:journals/corr/abs-1802-07584}.

Unfortunately, most of such methods were discovered on small dictionaries
and don't allow us to work in real sign language translation systems. In this
paper we are focused on building sign-level instead of a sentence-level
recognition model but with the ability to learn a good number of signs for
communication. In contrast to \cite{DBLP:journals/corr/abs-1812-01053} we
developed the model for continuous stream sign language recognition (instead of
clip-level recognition).

The main obstacle for gesture recognition (all the more so for translation)
system building is the limited amount of public datasets. The available datasets
are recorded with a minor number of signers and gestures, so the list of dataset
appropriate for training of deep networks datasets is mostly limited by
RWTH-PHOENIX-Weather \cite{forster-etal-2014-extensions} and MS-ASL
\cite{DBLP:journals/corr/abs-1812-01053}.

An insufficient amount of data causes over-fitting and limited model
robustness for changes in background, viewpoint, signer dialect. To overcome the
limitations of available databases, we reuse the best practices from
metric-learning area \cite{DBLP:journals/corr/abs-1812-05944}.

\begin{table*}[h]
\caption{Specification for S3D
MobileNet-V3-Large large backbone with residual spatio-temporal attention
modules.}
\label{table:backbone}
\centering
\begin{tabular}{c|c|c|c|c|c|c|c|c|c|c|c}
% \multicolumn{2}{c|}{Input size} & \multirow{2}{*}{Operator} & \multicolumn{2}{c|}{Kernel} & \multirow{2}{*}{\begin{tabular}[c]{@{}c@{}}Exp\\ size\end{tabular}} & \multirow{2}{*}{\begin{tabular}[c]{@{}c@{}}Num\\ out\end{tabular}} & \multirow{2}{*}{SE} & \multirow{2}{*}{NL} & \multicolumn{2}{c|}{Stride} & \multirow{2}{*}{Dropout} \\
\textbf{Sp. size} & \textbf{Temp. size} & \textbf{Operator} & \textbf{Sp. kernel} & \textbf{Temp. kernel} & \textbf{Exp size} & \textbf{Num out} & \textbf{SE} & \textbf{NL} & \textbf{Sp. stride} & \textbf{Temp. stride} & \textbf{Dropout} \\
\hline 
224 & 16 & conv3d & 3 & 1 & - & 16 & - & HS & 2 & 1 & - \\
112 & 16 & bneck & 3 & 5 & 16 & 16 & - & RE & 1 & 1 & \checkmark \\
112 & 16 & bneck & 3 & 3 & 64 & 24 & - & RE & 2 & 2 & \checkmark \\
56 & 8 & bneck & 3 & 3 & 72 & 24 & - & RE & 1 & 1 & \checkmark \\
56 & 8 & bneck & 5 & 3 & 72 & 40 & \checkmark & RE & 2 & 1 & \checkmark \\
28 & 8 & bneck & 5 & 3 & 120 & 40 & \checkmark & RE & 1 & 1 & \checkmark \\
28 & 8 & bneck & 5 & 5 & 120 & 40 & \checkmark & RE & 1 & 1 & \checkmark \\
28 & 8 & bneck & 3 & 5 & 240 & 80 & - & HS & 2 & 1 & \checkmark \\
14 & 8 & bneck & 3 & 3 & 200 & 80 & - & HS & 1 & 1 & \checkmark \\
14 & 8 & bneck & 3 & 3 & 184 & 80 & - & HS & 1 & 1 & \checkmark \\
14 & 8 & bneck & 3 & 5 & 184 & 80 & - & HS & 1 & 1 & \checkmark \\
14 & 8 & attention & 3 & 3 & - & 80 & - & - & 1 & 1 & - \\
14 & 8 & bneck & 3 & 3 & 480 & 112 & \checkmark & HS & 1 & 2 & \checkmark \\
14 & 4 & bneck & 3 & 3 & 672 & 112 & \checkmark & HS & 1 & 1 & \checkmark \\
14 & 4 & bneck & 5 & 3 & 672 & 160 & \checkmark & HS & 2 & 1 & \checkmark \\
7 & 4 & attention & 3 & 3 & - & 160 & - & - & 1 & 1 & - \\
7 & 4 & bneck & 5 & 3 & 960 & 160 & \checkmark & HS & 1 & 1 & \checkmark \\
7 & 4 & bneck & 5 & 3 & 960 & 160 & \checkmark & HS & 1 & 1 & \checkmark \\
7 & 4 & conv3d & 1 & 1 & - & 960 & - & HS & 1 & 1 & - \\
\end{tabular}
\end{table*}

\section{Methodology}

Our goal is to predict one of hand gestures
for each frame from the continuous input stream. To do that, we process the
fixed size sliding window of input frames. Experimentally, we've chosen to set
the number of input frames to 16 at constant frame-rate of 15. It captures,
roughly, 1 second of live video and covers the duration of the majority of ASL
gestures (according to the statistics of MS-ASL dataset). The extracted sequence
of frames is cropped according to the maximal (maximum is taken over all frames
in a sequence) bounding box of a person's face and both hands (only raised hands
are taken into account).  Finally, the cropped sequence is resized to 224 square
size producing a network input of shape $16 \times 224 \times 224$.

Unlike other solutions, we don't split network input into independent
streams for head and both hands
\cite{DBLP:journals/corr/abs-1906-02851}. Instead, we use a single RGB stream of
a cropped region that includes face and both hands of the signer to provide the
real-time performance.

You can find our demo application at Intel$^{\textregistered}$
OpenVINO\texttrademark OMZ\footnote{\url{https://github.com/opencv/open_model_zoo}}.
It employs a person detector, a tracker module and the ASL recognition
network itself along with all the necessary processing.

\subsection{Backbone design}
Instead of designing a custom lightweight
backbone adopted for inference on video stream we reuse a 2D backbone developed
for efficient computing at the edge. The logic behind this is based on the
assumption that the network efficient for 2D image processing will be a solid
starting point after extending it to additional temporal dimension due to high
correlation between the neighboring frames. We have selected MobileNet-V3
\cite{DBLP:journals/corr/abs-1905-02244} as a base architecture.

To extend a 2D backbone to 3D case, we follow the practices from the S3D
network\cite{DBLP:journals/corr/abs-1712-04851}: spatial and temporal separable
3D convolutions and top-heavy network design. According to the latter paradigm,
we remove temporal kernels from the very first convolution of a 3D backbone.

The default MobileNet-V3 bottleneck consists of three consecutive
convolutions: $1 \times 1$, depth-wise $k \times k$, $1 \times 1$. To convert it
into a 3D bottleneck following the concept of separable convolutions the last $1
\times 1$ convolution is replaced with a $t \times 1 \times 1$ one, where $t$ is
the temporal kernel size. In SE-blocks we carry out average pooling along
temporal dimension independently, so the shape of the attention mask is $T
\times 1 \times 1$, where $T$ is the temporal feature size. Following the
original MobileNet-V3 architecture we use different temporal kernels of sizes 3
and 5 but on contrasting positions.

Unlike spatial kernels, we don't use convolutions
with stride more than one for temporal kernels. To reduce the temporal size of a
feature map the temporal average pooling operator with appropriate kernel size
and stride sizes is used. Note, the positions of temporal pooling operations
are different from spatial ones.

One more change to the original MobileNet-V3 architecture is an addition of
two residual spatio-temporal attentions after the bottlenecks 9 and 12. See the
table \ref{table:backbone} for more details about the S3D MobileNet-V3 backbone
(the original table from the Mobilenet-V3 paper is supplemented by temporal
dimension-related columns).

\subsection{Spatio-Temporal Attention}

\begin{figure}[ht]
\centering
\includegraphics[width=0.45\textwidth]{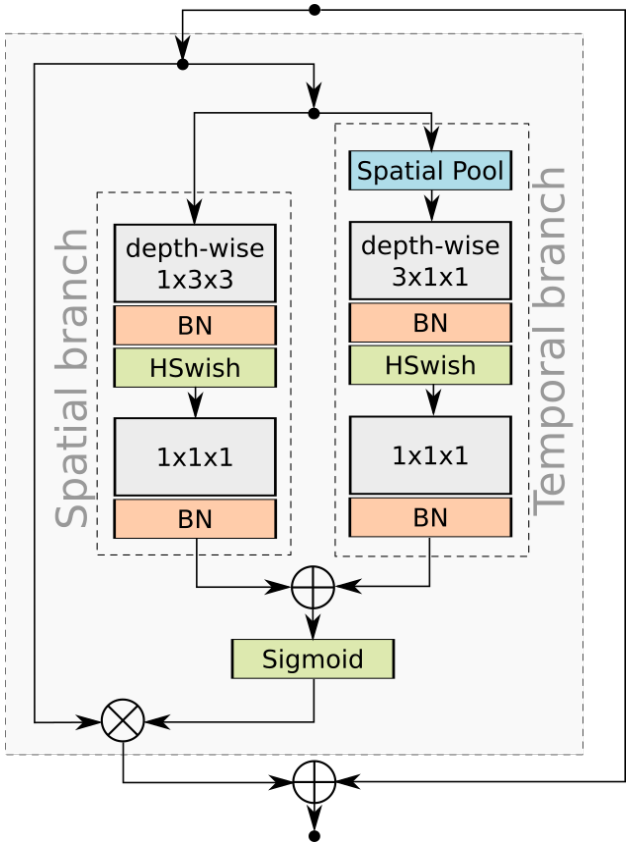}
\caption{Block-scheme of residual spatio-temporal attention module. "Spatial
Pool" block carries out global average pooling over spatial dimensions.}
\label{fig:attention_scheme}
\end{figure}

Inspired by \cite{DBLP:journals/corr/WangJQYLZWT17} and \cite{Dhingra_2019} we
reuse the paradigm of residual attention due to the possibility to insert it
inside the pre-trained network for training on a target task. One more advantage
is based on an ideology of consequence filtering of spatial appearance-irrelevant
regions and temporal motion-poor segments. Unlike the above solutions, we are
interested not only in unsupervised behavior of extra blocks but also in feature-level
self-supervised learning\footnote{Originally the term is related to the
unsupervised pre-training problem, like in \cite{DBLP:journals/corr/abs-1906-12340}
and \cite{DBLP:journals/corr/abs-1901-09005}.}.

To efficiently incorporate the attention module in 3D framework the
original single-stream block design is replaced by the two-stream design with
independent temporal and spatial branches. Each branch uses separable 3D
convolutions like in the bottleneck proposed above: consecutive depth-wise $1
\times 3 \times 3$ and $1 \times 1 \times 1$ convolutions with BN
\cite{DBLP:journals/corr/IoffeS15} and intermediate H-Swish activation function
\cite{DBLP:journals/corr/abs-1905-02244} for spatial stream (the only difference
for the temporal stream is in the first convolution which is depth-wise $3
\times 1 \times 1$). Then, both streams are added up and normalized by sigmoid
function during the inference stage (during the training stage the mask is
sampled -- see next section). For more details see Figure \ref{fig:attention_scheme}.

\begin{figure}[ht]
\centering
\includegraphics[width=0.45\textwidth]{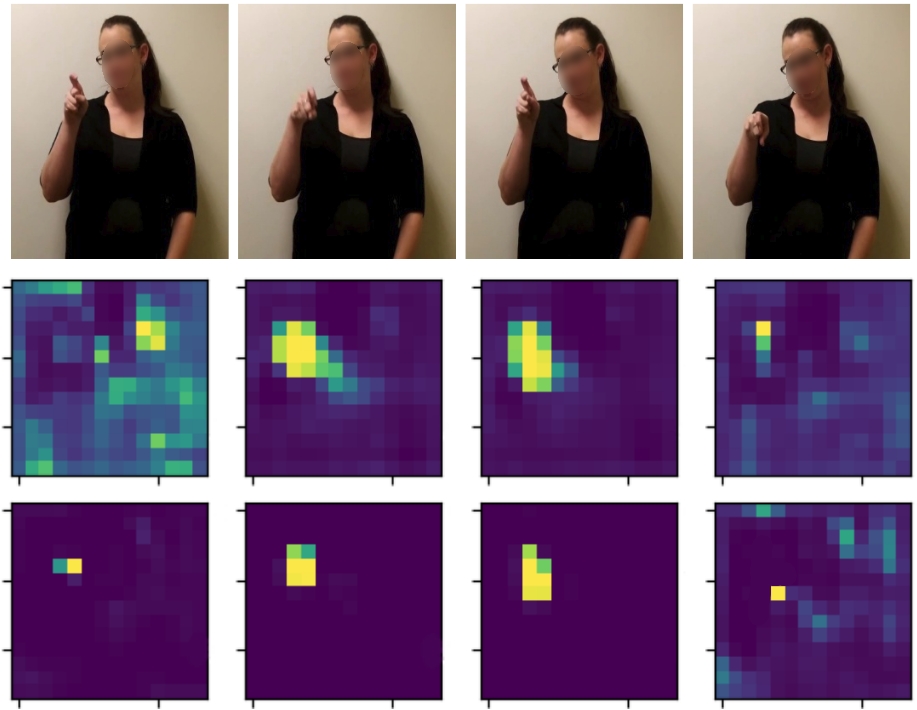}
\caption{Example of spatio-temporal attention masks. In rows from top to bottom:
the original sequence of RGB frames, corresponding attention maps without
auxiliary loss and attention maps after training with the proposed
self-supervised loss.}
\label{fig:attention_masks}
\end{figure}

The main drawback of using an attention module in unsupervised manner is a
weak discriminative ability of learnt features (take a look on Figure
\ref{fig:attention_masks}, where attention masks from the second row are too noisy to
extract robust features). As a result, even attention-augmented networks cannot
fix an incorrect prediction and no significant benefit from using attention
mechanisms can be observed. In our opinion, it's because no extra information is
provided during training to force the network to fix the prediction by focusing
on the most relevant spatio-temporal regions rather than soft tuning over all
model parameters (some kind of the "Divide and Conquer" principle).

We propose to encourage the spatio-temporal homogeneity by using the total
variation (TV) loss \cite{DBLP:journals/corr/MahendranV14} over the
spatio-temporal confidences. In addition, to force the attention mask to be
sharp, the TV-loss is modified to work with hard targets ($0$ and $1$ values):

\begin{equation}
\label{eq:tvloss}
L_{TV} = \frac{1}{TMN} \sum_{t,i,j} |s_{tij} - I(\frac{1}{|N_{tij}|} \sum_{n \in N_{tij}} s_n > 0.5)|
\end{equation}

\noindent where $s_{tij}$ is a confidence score at a spatial position $i,j$ and
a temporal position $t$ of a spatio-temporal confidence map of shape $T \times M
\times N$, $N_{tij}$ is a set of neighboring spatio-temporal positions of
element $s_{tij}$ and $I(\cdot)$ is an indicator function. Note, we use TV-loss
over spatio-temporal confidences, rather than logits.

Another drawback of attention modules is a tendency of getting stuck in
local minima (e.g. a network can learn to mask a central image region only
regardless of input features). To overcome the above problem we propose to learn
the distribution of masks and sample one during training\footnote{The idea is
  related to energy-based learning, like in
  \cite{turner2018metropolishastings}.} and use the expected value during
inference. For this purposes, we reuse the Gumbel-Softmax trick
\cite{jang2016categorical}, but for sigmoid function
\cite{DBLP:journals/corr/abs-1805-02336}. As you can see on figure
\ref{fig:attention_masks}, the proposed methods allow us to train a much sharper and
robust attention mask.

\subsection{Metric-Learning approach}
The default approach to train an action
recognition network is to use Cross-Entropy classification loss. This method
works fine for large size datasets and there is no reason to change
it. Unfortunately, if we are limited in available data or the data is
significantly imbalanced, then sophisticated losses are needed.

We are inspired by the success of metric-leaning approach to train networks
on the limited size datasets to solve the person re-identification problem. So,
we follow the practice to use the AM-Softmax
\cite{DBLP:journals/corr/abs-1801-05599} loss \footnote{Originally the loss has
been designed for the Face Verification problem but has become the standard
for several adjacent tasks.}  with some auxiliary losses to form the manifold
structure according the view of ideal geometrical structure of such space.
Similar to \cite{DBLP:journals/corr/abs-1905-00292} we replace constant scale
for logits by the straightforward schedule: gradual descent from 30 to 5 during
40 epochs.

Additionally, to prevent over-fitting on the simplest samples we follow the
technique proposed in \cite{DBLP:journals/corr/abs-1906-04692} to regularize the
cross-entropy loss by addition of max-entropy term:

\begin{equation}
\label{eq:am_softmax_max_entropy}
L_{AM} = [L_{cross}(p) - \alpha H(p)]_{+}
\end{equation}

\noindent where $p$ is the predicted distribution and $H(\cdot)$ is the entropy
of input distribution.

As mentioned in \cite{DBLP:journals/corr/abs-1812-02465}, AM-Softmax loss
forms a global structure of manifold but the decision boundary of exact classes
is also defined by a local interaction between neighboring samples. So, the
PushPlus $L_{push}$ loss between samples of different classes in batch is used,
too. In a similar manner, the push loss is introduced between the centers of
classes to prevent the collapse of close clusters (aka $L_{cpush}$ loss). The
final loss is a sum of all of the mentioned above losses: $L = L_{AM} + L_{push}
+ L_{cpush}$.

Further, metric-learning approach allows us to train networks that are
close to large networks in terms of quality, but are much lighter and, thereby,
faster \cite{DBLP:journals/corr/abs-1812-02465}.

\subsection{Network Architecture}
The sign gesture recognition network
architecture consists of S3D MobileNet-V3 backbone, reduction spatio-temporal
module and classification metric-learning based head. The backbone outputs the
feature map of size $4 \times 7 \times 7$ (the number of channels is unchanged
for MobileNet-V3 and equals to 960) thereby reducing input by 32 times in
spatial dimension and 4 times in temporal one. Then, the spatio-temporal module
carries out reduction of the final feature map by applying global average
pooling. Lastly, the obtained vector is convolved with $1 \times 1 \times 1$
kernel to align the number of channels to the target of 256 (we have
experimented with different embedding sizes but the best trade off between speed
and accuracy is obtained with that value). The output embedding vector is $L_2$
normalized. Also, we use BN layer before the normalization stage, like in
\cite{DBLP:journals/corr/abs-1807-11042} and
\cite{DBLP:journals/corr/abs-1906-08332}.

\subsection{Model training}
\label{chap:training}
To enhance the situation with model robustness
against appearance cluttering and motion shift, a number of image- and
video-level augmentation techniques is used: brightness, contrast, saturation
and hue image augmentations, plus, random crop erasing
\cite{DBLP:journals/corr/abs-1708-04896} and the mixup
\cite{DBLP:journals/corr/abs-1710-09412} (with a random image from ImageNet
\cite{ILSVRC15} and a gesture clip without mixing the labels).  All the
mentioned augmentations are sampled once per clip and applied for each frame in
the clip identically. Additionally, to force the model to guess about action of
the partially presented sequence of sign gesture we use the temporal jitter for
temporal limits of action.  During training we set the minimal intersection
between ground-truth and augmented temporal limits to 0.6.

Additionally, to prevent from over-fitting, we augment training at the
network level by addition of continuous dropout \cite{shen2019continuous} layer
inside each bottleneck (instead of single one on top of the network) as it was
originally proposed in \cite{DBLP:journals/corr/PaszkeCKC16}.

The network training procedure cannot converge when
starting from scratch. So, we use the two-stage pre-training scheme: on the first
stage the 2D Mobilenet-V3 backbone is trained on ImageNet \cite{ILSVRC15}
dataset. Then the S3D MobileNet-V3 network equipped with residual
spatio-temporal attention modules and metric-learning losses is trained on
Kinetics-700 \cite{DBLP:journals/corr/abs-1907-06987} dataset. The only change
before starting the main training stage is replacing the centers of classes (the
weight matrix with which an embedding vector should be multiplied) to randomly
picked ones according to the configuration of MS-ASL dataset with 1000 classes
(unlike the mentioned paper with didn't see the benefit from training directly
on 100 classes due to fast over-fitting).

The final network has been trained on two GPUs by 14 clips per node with
SGD optimizer and weight decay regularization using PyTorch framework
\cite{paszke2019pytorch}. The initial learning rate is set to $0.01$ with single
drop after the 25th epoch. Additionally, warm-up
\cite{DBLP:journals/corr/abs-1810-13243} is used during the first 5 epochs
starting from $0.0001$ learning rate and PR-Product (for last inner product
only) \cite{DBLP:journals/corr/abs-1904-13148} to enable parameter tuning around
the convergence point. Note, due to significant over-fitting we use early
stopping after approximately 30 epochs.

\section{Experiments}

\subsection{Data}
Sign language databases and American Sign
Language (ASL), in particular, are hard to collect due to the need of capable
signers. The first attempt to build a large-scale database has been made by
\cite{athitsos2008american} when they published ASLLBD database. We have
experimented with this dataset but the final model suffers from significant
domain shift and doesn't allow us to run it on a video with an arbitrary signer
or cluttered background, even though it achieves nearly maximal quality on the
train-val split. It's because the database has been collected with a limited
number of signers (less then ten) and constant background. So, for the
appearance-based solutions the emphasized database is not very useful.

The major leap has been made when MS-ASL
\cite{DBLP:journals/corr/abs-1812-01053} dataset has been published. It includes
more than 25000 clips over 222 signers and covers 1000 most frequently used ASL
gestures.  Additionally, the dataset has a predefined split on train, val and
test subsets. To utilize the maximal number of lacking samples of sign gestures,
we train the network on full 1000-class train subset, but our goal is high
metrics on the 100-class subset. Unlike the previously mentioned paper, we
didn't see the benefit of using 100-class subset directly for
training. Moreover, we have observed significant over-fitting even for the much
smaller network in comparison with the I3D baseline from the paper. Nonetheless,
we use MS-ASL dataset to train and validate the proposed ASL recognition model.

Note, the paper proposes to test models (and provides baselines) for MS-ASL
dataset under the clip-level setup. It implies the knowledge about the time of
start and end of the sign gesture sequence. In this paper, we are focused on
developing continuous stream action recognition model which should work on the
unaligned (unknown start and end) sequence of sign gesture. So, the baselines
from the original paper and the current paper are not directly comparable due to
a more complicated scenario that we consider (we hope the future models will be
compared under more suitable continuous recognition scenario).

\subsection{Test protocol}

To better model the scenario of action
recognition of a continuous video stream, we follow the next testing
protocol. From each sequence of annotated sign gestures we select the central
temporal segment with length equal to the network input (if the length of the
sequence is less than the network input then the first frame is duplicated as
many times as required). After that, the sequence of frames is cropped according
to the mean bounding box of person (it includes head and two hands of a
signer). The predicted score on this sequence is considered a prediction for the
input sample (no over-sampling or other test-time techniques for metric boosting
are used).

We measure mean top-1 accuracy and mAP metrics. Unlike the original MS-ASL
paper we don't use top-5 metric to level the annotation noise in the dataset
(incorrect labels, mismatched temporal limits) due to weak correlation between
the model robustness and high value of this metric (our experiments showed that
a model with high top-5 metric can demonstrate low robustness in live-mode
scenario).

Note, as mentioned in the original paper
\cite{DBLP:journals/corr/abs-1812-01053}, the data includes significant noise in
annotation. Likewise, we observed many mismatches in annotated sign gestures, so
it's expected that the real model performance is higher than the metric values
suggest and it was confirmed indirectly by the impressive model accuracy in live
mode.

\subsection{Ablation study}
Here, we present the ablation study (see the
table \ref{table:abstudy}). The baseline model includes training in continuous
scenario with default AM-Softmax loss and scheduled scale for logits. As you can
see from the table, the first solution is much lower than the best one due to
weak learnt features even though it uses metric-learning approach from the very
beginning.

\begin{table}[h]
\caption{Ablation study on MS-ASL test-100 dataset}
\label{table:abstudy}
\centering
\begin{tabular}{l|cc}
%\multicolumn{1}{c|}{\multirow{2}{*}{\textbf{Method}}} & \multicolumn{2}{c}{\textbf{MS ASL 100}} \\
%\multicolumn{1}{c|}{} & \textbf{top-1} & \textbf{mAP} \\
\textbf{Method} & \textbf{top-1} & \textbf{mAP} \\
\hline 
AM-Softmax & 73.93 & 76.01 \\
+ temporal jitter & 76.49 & 78.63 \\
+ dropout in each block & 77.11 & 79.45 \\
+ continuous dropout & 77.35 & 79.85 \\
+ extra ml losses & 80.75 & 82.34 \\
+ pr-product & 80.77 & 82.67 \\
+ mixup & 82.61 & 86.40 \\
+ spatio-temporal attention & 83.20 & 87.54 \\
+ hard TV-loss & 85.00 & 87.79 \\
\end{tabular}
\end{table}

The first thing that should be fixed is weak annotation that includes
mostly incorrect temporal segmentation of gestures. To fix it we let loose the
condition to match the ground-truth temporal segment and a network input. The
proposed change improves both metrics with a decent gap.

Then, the issue with insufficiently large and diverse dataset should be
handled. To do that, we follow the practice of using dropout regularization
inside each bottleneck. At the expense of reduction of a model capacity, the
model enhances collective decision making \cite{JMLR:v15:srivastava14a} by
suppression of some kind of "grandmother cell" \cite{gross2002genealogy}. One
more small step is to replace the default Bernoulli distribution with continuous
Gaussian distribution, like in \cite{shen2019continuous}. After this, the model
enhances both metrics but still suffers from domain shift problem
\cite{DBLP:journals/corr/abs-1809-09478}.

To overcome the mentioned above issue we have proposed to go deeper into
metric-leaning solutions by introducing local structure losses
\cite{DBLP:journals/corr/abs-1812-02465}. As you can see, it allows us to score
higher than 80 percent for both metrics. Additionally, the PR-Product is used to
force learning near zero-gradient regions. Note, in our experiments the usage of
PR-Product was justified with extra metric-learning losses only.

Another improvement is tied to increasing the variety of appearance by
mixing video clips with random images (see the description of the implemented
mixup-like augmentation in \ref{chap:training}). The amount of the accuracy
increase tells us about the importance of appearance diversity for neural
network training.

The last leap is provided by using the residual spatio-temporal attention
module with the proposed self-supervised loss.

\subsection{Results}
As it was mentioned earlier, we cannot compare
the proposed solution with a previous one on MS-ASL dataset because we have
changed the testing protocol from the clip-level to continuous-stream
paradigm. The final metrics on MS-ASL dataset (test split) are presented in
table \ref{table:result_metrics}.  Note, as mentioned in the Data section, the
quality of the provided annotation doesn't allow us to measure the real power of
the trained network even after manual filtering of the data (we carried out
simple filtering to exclude empty or incorrectly cut gesture sequences).

\begin{table}[h]
\caption{Results of continuous ASL recognition model on MS-ASL dataset}
\label{table:result_metrics}
\centering
\begin{tabular}{l|ll}
\textbf{MS ASL split} & \textbf{top-1} & \textbf{mAP} \\
\hline
100 signs & 85.00 & 87.79 \\
200 signs & 79.66 & 83.06 \\
500 signs & 63.36 & 70.01 \\
1000 signs & 45.65 & 55.58 \\
\end{tabular}
\end{table}

The final model takes 16 frames of $224 \times 224$ image size as input at
the constant 15 frame-rate and outputs embedding vector of 256 floats. As far as
we know, the proposed solution is the fastest ASL Recognition model (according
our measurements on Intel$^{\textregistered}$ CPU) with competitive metric values
on MS-ASL dataset. The model has only 4.13 MParams and 6.65 GFlops. For more
details see table \ref{table:result_params}.

\begin{table}[h]
\caption{Continuous ASL recognition model specification}
\label{table:result_params}
\centering
\begin{tabular}{c|c|c|c|c}
%\multicolumn{2}{c|}{Input size} & \multirow{2}{*}{\begin{tabular}[c]{@{}c@{}}Embd\\ size\end{tabular}} & \multirow{2}{*}{MParams} & \multirow{2}{*}{GFlops} \\
\textbf{Spatial size} & \textbf{Temporal size} & \textbf{Embd size} & \textbf{MParams} & \textbf{GFlops} \\
\hline
$224 \times 224$ & 16 & 256 & 4.13 & 6.65 \\
\end{tabular}
\end{table}

\subsection{Discussion}
Presently, graph-based approaches
\cite{materzynska2019somethingelse}, \cite{chen2020general},
\cite{liang2020visualsemantic} gain popularity for action recognition tasks.
Aforementioned methods rely on modeling the interactions between objects in a
frame through time. It looks like the idea from \cite{zhang2019temporal} can be
transferred to gesture recognition challenge but, on practice, the addition of
residual attention modules with simple global average pooling reduction operator
shows similar quality without the need of extra computation. In our opinion, the
most appropriate explanation of the mentioned behavior is that a sign gesture
has a fixed spatial (placement of two hands and face) and temporal (transition
of fingers through time) structure which can be easily captured by 3D neural
network with sufficient spatio-temporal receptive field. However, incorporating
low-level design of graph-based approach for feature extractor directly could
give a fresh view on the proposed solution and we hope it will be done in the
future.

\section{Conclusion}
In this paper we propose the lightweight ASL
gesture recognition model which is trained under the metric-learning framework
and allows us to recognize ASL signs in a live stream. Besides that, for better
model robustness to appearance changes, it's proposed to use residual
spatio-temporal attention with the auxiliary self-supervised loss. The results
show that the proposed gesture recognition model can be used in a real use case
for ASL sign recognition.

\bibliographystyle{IEEEtran}
% argument is your BibTeX string definitions and bibliography database(s)
\bibliography{IEEEabrv,egbib}
%
% <OR> manually copy in the resultant .bbl file
% set second argument of \begin to the number of references
% (used to reserve space for the reference number labels box)
% \begin{thebibliography}{1}
%
% \bibitem{IEEEhowto:kopka}
% H.~Kopka and P.~W. Daly, \emph{A Guide to \LaTeX}, 3rd~ed.\hskip 1em plus
%   0.5em minus 0.4em\relax Harlow, England: Addison-Wesley, 1999.
%
% \end{thebibliography}

% that's all folks
\end{document}